\documentclass[10pt,journal,compsoc]{IEEEtran}
\usepackage[center]{caption}

\usepackage{caption}
\usepackage{subcaption}

\usepackage{multirow}
\usepackage{booktabs}
\usepackage{listings}
\usepackage{xcolor,colortbl}

\usepackage{amsmath}
\usepackage{algorithm}
\usepackage[noend]{algpseudocode}

\usepackage{hyperref}

\definecolor{Gray}{gray}{0.85}
\newcolumntype{g}{>{\columncolor{Gray}}c}

\newlength{\Oldarrayrulewidth}
\newcommand{\Cline}[2]{%
  \noalign{\global\setlength{\Oldarrayrulewidth}{\arrayrulewidth}}%
  \noalign{\global\setlength{\arrayrulewidth}{#1}}\cline{#2}%
  \noalign{\global\setlength{\arrayrulewidth}{\Oldarrayrulewidth}}}

\ifCLASSOPTIONcompsoc
  \usepackage[nocompress]{cite}
\else
  \usepackage{cite}
\fi
\ifCLASSINFOpdf
\else
\fi
\usepackage{array}

\usepackage{adjustbox}
\usepackage{enumitem}

\begin{document}

\title{IMAGO: A family photo album dataset for a socio-historical analysis of the twentieth century}

\author{Lorenzo~Stacchio,
        Alessia~Angeli,
        Giuseppe~Lisanti,
        Daniela~Calanca,
        Gustavo~Marfia

\IEEEcompsocitemizethanks{
\IEEEcompsocthanksitem This work has been submitted to the IEEE for possible publication. Copyright may be transferred without notice, after which this version may no longer be accessible.\protect\\
\IEEEcompsocthanksitem L. Stacchio is with the Department for Life Quality Studies, University of Bologna, Bologna,
Italy.\protect\\
E-mail: lorenzo.stacchio2@unibo.it
\IEEEcompsocthanksitem A. Angeli is with the Department
of Computer Science and Engineering, University of Bologna, Bologna,
Italy.\protect\\
E-mail: alessia.angeli2@unibo.it
\IEEEcompsocthanksitem G. Lisanti is with the Department
of Computer Science and Engineering, University of Bologna, Bologna,
Italy.\protect\\
E-mail: giuseppe.lisanti@unibo.it
\IEEEcompsocthanksitem D. Calanca is with the Department for Life Quality Studies, University of Bologna, Bologna,
Italy.\protect\\
E-mail: daniela.calanca@unibo.it
\IEEEcompsocthanksitem G. Marfia is with the Department for Life Quality Studies, University of Bologna, Bologna,
Italy.\protect\\
E-mail: gustavo.marfia@unibo.it
}

\thanks{}}


\IEEEtitleabstractindextext{%
\begin{abstract}
Although one of the most popular practices in photography since the end of the 19th century, an increase in scholarly interest in family photo albums dates back to the early 1980s. Such collections of photos may reveal sociological and historical insights regarding specific cultures and times. They are, however, in most cases scattered among private homes and only available on paper or photographic film, thus making their analysis by academics such as historians, social-cultural anthropologists and cultural theorists very cumbersome. In this paper, we analyze the IMAGO dataset including photos belonging to family albums assembled at the University of Bologna's Rimini campus since 2004. Following a deep learning-based approach, the IMAGO dataset has offered the opportunity of experimenting with photos taken between year 1845 and year 2009, with the goals of assessing the dates and the socio-historical contexts of the images, without use of any other sources of information. Exceeding our initial expectations, such analysis has revealed its merit not only in terms of the performance of the approach adopted in this work, but also in terms of the foreseeable implications and use for the benefit of socio-historical research. To the best of our knowledge, this is the first work that moves along this path in literature.
\end{abstract}

}

\maketitle

\IEEEdisplaynontitleabstractindextext

\IEEEpeerreviewmaketitle

\IEEEraisesectionheading{\section{Introduction}
\label{sec:introduction}}

\IEEEPARstart{F}{ollowing} Kodak's invention of the first megapixel sensor in 1986, digital photography has slowly grown to substitute its analog predecessor, playing a key role in the early 21st century digital revolution and social transformation \cite{peres2014concise, serafinelli2018digital}. As a relevant example, photography has modified the way mobile phones are used, as their integration of digital cameras has at once fostered an exponential growth of the photos that are shot and uploaded to the Internet every year, as well as a paradigm shift in mobile communications, which today rely on high quality multimedia \cite{yin2013socialized,Big_data_infrastructure_I,borcoci2010novel,rainer2016statistically}. These phenomena have proven to be game-changers for both how people communicate and the bloom of new fields of research, as both academia and industry have put to good use such plethora of visual data to train, develop and apply Artificial Intelligence (AI) models to a variety of different problems (e.g., face recognition, autonomous driving) \cite{lemley2017deep,li2019deep, chen2020learning,vaccaro2020image, 2020potential}.

Now, while a wealth of research is being devoted to the processing and analysis of digital images, much has to be done regarding analog ones, mainly because printed images representing a place (or an environment) at a given time may be: (i) scattered in numerous public and private collections, (ii) of variable quality, and, (iii) damaged due to hard or continued use or exposure. In addition, any analysis by means of image processing and computer vision algorithms requires the cumbersome and potentially quality degrading initial digitization step.

However, despite the complications and challenges brought on by analog photographs, they nevertheless represent an unparalleled source of information regarding the recent past: in fact, no other visual media has been used as pervasively to capture the world throughout the 20th century, as the availability of consumer grade photo cameras supported the spread and popularity of vernacular photography practices (e.g., travel photos, family snapshots, photos of friends and classes) \cite{mitman2016documenting,vernacular_moma}.

Family photo albums represent an example of vernacular photography that has drawn the attention of researchers and public institutions. A recent work defines family photo albums \textit{a globally circulating form that not only takes locally specific forms but also ``produces localities'' that creates and negotiate individual stories} \cite{sandbye2014looking}. Along the same lines, in another relevant contribution, family albums \textit{represent a reference point for the conservation, transmission and development of a community Social Heritage} \cite{calanca2011italians}. In essence, scholars from different fields agree in identifying such type of photography collections as capable of capturing salient features regarding the evolution of local communities in space and time.

However, a large-scale analysis of such collections of photos is often impeded by their numbers, as it would be exceedingly burdensome for historians and anthropologists to manually verify the characteristics of more than a few hundreds, considering also that in many cases no associated descriptions are available. This is why contributions in this field, normally, base their findings on the study of small corpora of photos~\cite{sandbye2014looking, calanca2011italians}.

\begin{figure}[t]
\captionsetup[subfigure]{labelformat=empty}
   \centering
   \begin{subfigure}{0.48\linewidth}
     \centering
     \includegraphics[width=\linewidth]{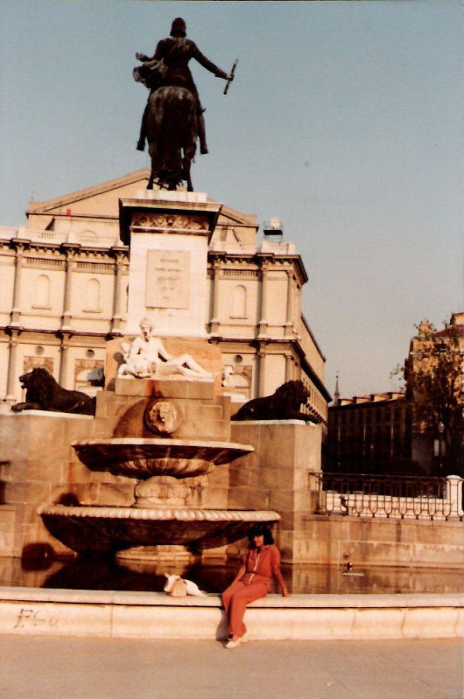}
     \caption{Free-time}
   \end{subfigure}
   \hfill
   \begin{subfigure}{0.48\linewidth}
     \centering
     \includegraphics[width=\linewidth]{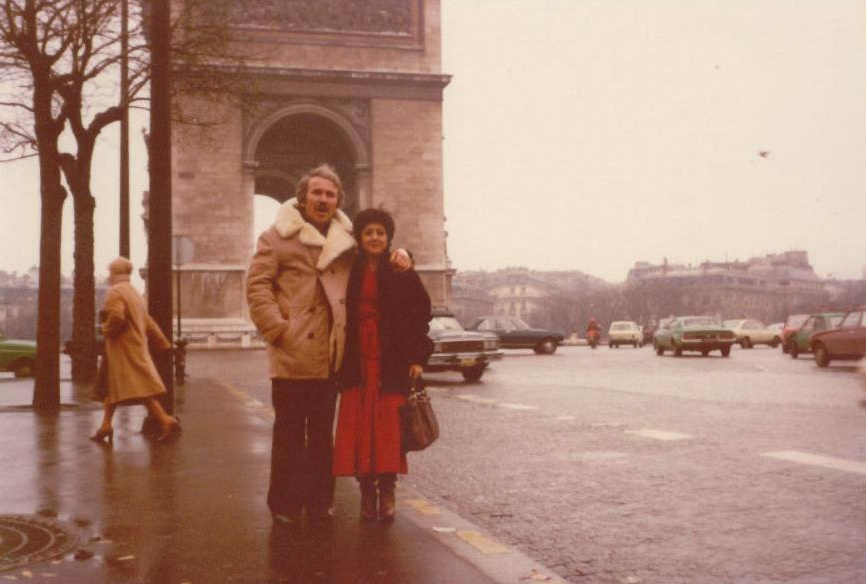}
     \caption{Fashion}
   \end{subfigure}
\caption{Similarity between pictures labelled as belonging to different socio-historical classes.}
\label{fig_sim_semantic_fight}
\end{figure}

This work addresses such problem, as it focuses on learning how to recognize a set of salient features of socio-historical interest exploiting the IMAGO collection of family photos started in year 2004 at the Rimini Campus of the University of Bologna~\cite{calanca2011italians}.
In addition, this work represents a first contribution for the following:

\begin{itemize}
    \item The introduction of a ``Family Album'' collection comprising over 80,000 photos taken between 1845 and 2009, belonging to ca 1,500 families, primarily from the Emilia-Romagna and immediately neighboring regions in Italy. 16,642 of such images have been labelled;
    \item The investigation of a new classification problem, to the best of our knowledge, regarding the identification of the socio-historical context of an image (according to classes individuated in socio-historical literature). In this case a model, given a photo, is requested to provide the socio-historical context of its subject. This amounts to quite a complicated task, naturally prone to ambiguities, as it can be easily seen from Fig.~\ref{fig_sim_semantic_fight} (Free-Time and Fashion, for example, share some common tracts, making it difficult to correctly discriminate between them);
    \item A thorough analysis using different deep learning architectures for both the image dating and the socio-historical context classification task. For photography and socio-historical scholars it is important to underline that the results of such analysis have been obtained with the sole use of the images at hand, without having the opportunity of resorting to external and additional sources of information.
\end{itemize}

The rest of the paper is organized as follows: in Section~\ref{sec:related-work} we review the state of the art that falls closest to this contribution. Section~\ref{sec:imago} provides a description and the main characteristics of the IMAGO dataset. Section~\ref{sec:model} and Section~\ref{sec:results} present and validate several deep neural networks models applied to the proposed dataset, in order to highlight the difficulties and define an evaluation baseline. Later, in Section~\ref{sec:discussion} a review of the patterns learned by the deep learning models is provided. Finally, in Section~\ref{sec:conclusions} an overall discussion is carried out, along with possible directions of future works.

\section{Related Work}
\label{sec:related-work}

Only a few works have proposed so far the analysis of collections of vernacular photographs, also taking into account analog ones~\cite{fernando2014color,ginosar2015century,salem2016analyzing,muller2017picture}. The attention of such contributions has focused mostly on the dating of the images inside the collections. In the following, a discussion is offered regarding how such task was implemented.

In~\cite{fernando2014color}, the authors addressed the problem of dating color photos taken between 1930 and 1970, belonging to a combination of vernacular and landscape photography, considering a dataset of 1,325 photos. To this aim, they verified whether specific characteristics of the device used to acquire the photographs, as the distribution of derivatives and the angles drawn by three consecutive pixels in the RGB space, could be put to good use to date a photograph. 
An accuracy of 85.5\% was achieved for the decades considered in the classification task. The proposed solution, though, is device dependent and may return inaccurate results when pictures are shot with particularly old cameras.

Another work employed a deep learning approach to analyze and date 37,921 historical frontal-facing American high school yearbook photos taken from 1928 to 2010~\cite{ginosar2015century}. Here, a Convolutional Neural Network (CNN) architecture was built to analyze people's faces and predict the year in which a photo was taken. The author observed the performance of the dating task to be gender-dependent: (i) considering the ability to individuate the exact year, a 10.9\% accuracy was obtained with women and 5.5\% with men, while (ii) 65.3\% and 46.4\% values, for women and men, respectively, were obtained accepting a $\pm$ 5 years tolerance.

Along the same lines, also the authors of~\cite{salem2016analyzing} presented a dataset containing images of students taken from high school yearbooks, covering in this case the 1950 to 2014 time span (considering 1,400 photos per year). Their results confirm that human appearance is strongly related to time. Moreover, they also resorted to CNNs to estimate when an image was captured, in order to evaluate the quality of color vs grayscale images containing the following features: (i) faces, (ii) torsos (i.e., upper bodies including people's faces), and, (iii) random regions of images. The best performance was obtained considering color images portraying the torso of people. Thanks to the size and balance in time of the dataset and to the fact that it represents the torso of people in homogeneous environments, the authors were able to obtain an accuracy of 41.6\% for the exact year estimation problem and a 75.8\% for the $\pm$ 5 years one. 

In~\cite{muller2017picture}, instead, the dating task was implemented through the analysis of images belonging to years 1930 through 1999. As for~\cite{fernando2014color}, vernacular and landscape photos are considered, amounting to at most 25,000 pictures per year. The authors proposed different baselines relying on deep CNNs, considering the dating as both a regression and a classification task. With this work the authors were able to reach fair results, precisely 51.8\% on the decades dating task, with heterogeneous types of photos.

The first four rows of Table~\ref{table:dataset_information_comparison} report the main characteristics (image content, number of images and covered time span) of the original photo archives employed in this Section (in most cases only specific selections of such archives have been analyzed by means of image processing techniques). To provide an easy to see comparison, the same information, regarding the IMAGO collection (i.e., the collection originating the dataset analyzed in this work), is provided in the last row of the same Table.

\begin{table}[h!]\large
\renewcommand{\arraystretch}{1.6}
\center
\begin{adjustbox}{width=\linewidth}
\begin{tabular}{|l|l|l|l|l|}
\hline
\textbf{Original dataset} & \textbf{Characteristics} & \textbf{Cardinality} & \textbf{Period} \\
\hline
\cite[B. Fernando et al.]{fernando2014color} &  \vtop{\hbox{\strut vernacular and}\hbox{\strut landscape photos}} & 1,325 & 1930 - 1979 \\
\hline
\cite[S. Ginosar et al.]{ginosar2015century} & \vtop{\hbox{\strut high school}\hbox{\strut yearbooks portraits}} & 168,055
& 1905 - 2013
\\
\hline
\cite[T. Salem et al.]{salem2016analyzing} & \vtop{\hbox{\strut high school}\hbox{\strut yearbooks portraits}} & ca 600,000
& 1912 - 2014
\\
\hline
\cite[E. M\"{u}ller et al.]{muller2017picture} & \vtop{\hbox{\strut vernacular and}\hbox{\strut landscape photos}} & 1,029,710 & 1930 - 1999 \\
\hline
 \textbf{IMAGO collection} & \vtop{\hbox{\strut \textbf{family album}}\hbox{\strut \textbf{analog photos}}} & \textbf{ca 80,000} & \textbf{1845 - 2009} \\
\hline
\end{tabular}
\end{adjustbox}
\caption{Original dataset comparison.}
\label{table:dataset_information_comparison}
\end{table}

It is now important to highlight that all of the contributions discussed in this Section focused on the dating of vernacular photographs shot in heterogeneous settings. None, however, to the best of our knowledge, has considered implementing the classification of such kind of photos according to their context. In this respect, the contribution presented in this paper not only performs the dating of single images, but also analyzes the socio-historical context of the photos belonging to family albums. Nevertheless, for the dating task, we also combined three different features, extracted from the same photo (the whole image along with the faces and the people's full figures it contains) and observed that the accuracy of the proposed models increases as more people are represented in a photo. Finally, the dataset analyzed in this work is the only one solely composed by analog photos.

\begin{figure*}[t]
   \centering
   \begin{subfigure}{0.45\linewidth}
     \centering
     \includegraphics[width=\linewidth]{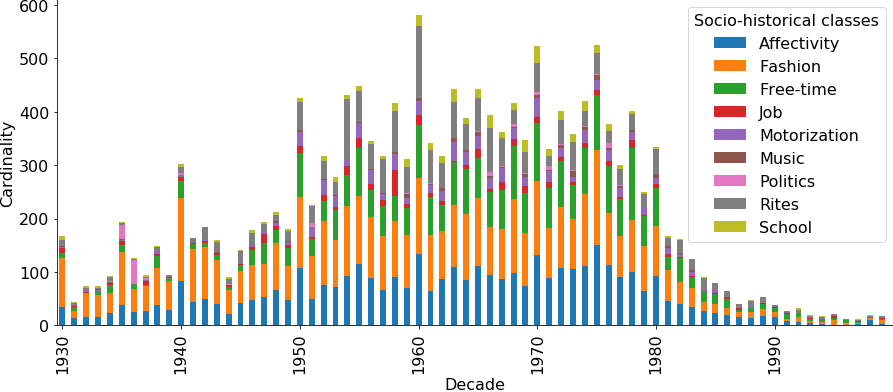}
     \caption{Classes distribution}
     \label{fig:distr_imago_normal}
   \end{subfigure}
   \hfill
   \begin{subfigure}{0.5\linewidth}
     \centering
     \includegraphics[width=\linewidth]{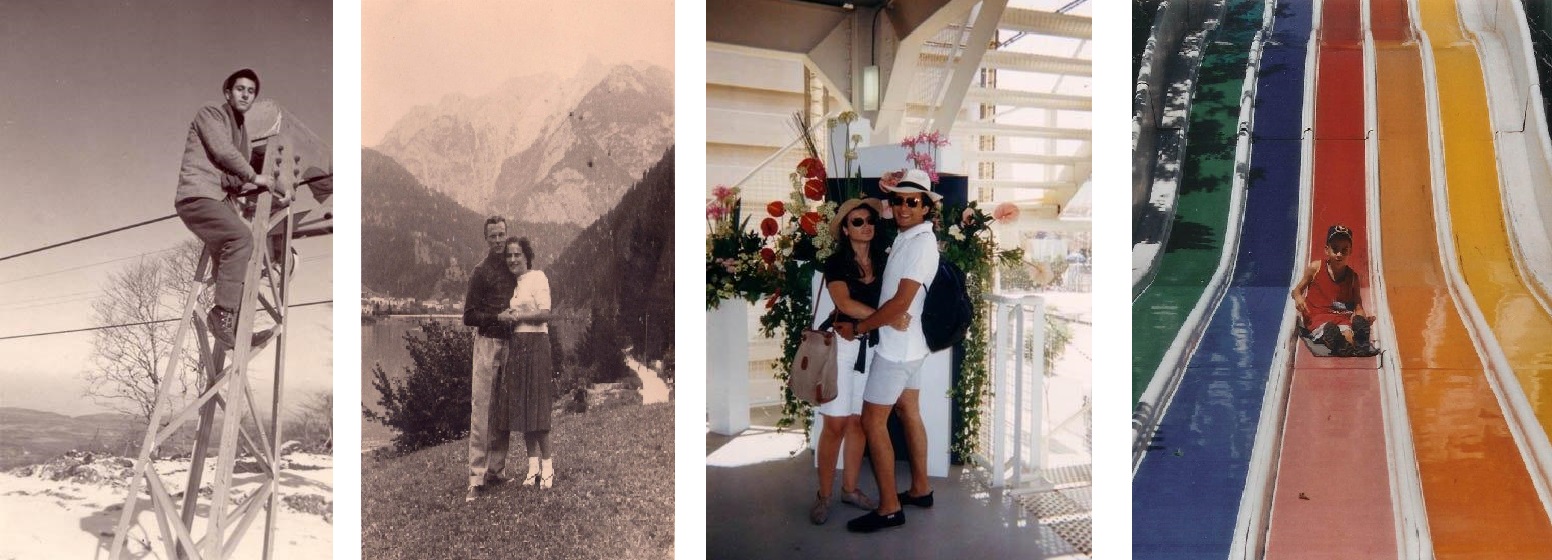}
     \caption{Sample images}
     \label{fig:samples_imago}
   \end{subfigure}
   \caption{IMAGO characteristics.}
\end{figure*}

\section{IMAGO dataset}
\label{sec:imago}

The IMAGO project is a digital collection of analog family photos year by year gathered and conserved by the Department for Life Quality Studies of the University of Bologna (a general overview may be found at 
\href{http://imago.unibo.it}{\texttt{imago.unibo.it}}). On the whole, it comprises ca 80,000 photos, taken between 1845 and 2009, belonging to ca 1,500 family albums. 16,642 images have been labeled by the bachelor students in ``Fashion Cultures and Practices'' course, under the supervision of the socio-historical faculty. The metadata associated to each image includes its shooting year, a short textual description, the location (city and nation) where it was taken and a unique socio-historical label describing the category a scene belongs to~\cite{calanca2011italians}. The aforementioned subset of the IMAGO collection composed by 16,642 labelled images was considered as our dataset, namely the IMAGO dataset\footnote{The IMAGO resources, comprehending the dataset and the models, are available upon request.}.
From now on, we will consider the IMAGO dataset focusing on the individuation of the year and the socio-historical category (as described by the associated label). To this aim, the fourteen socio-historical categories, which have been individuated in~\cite{imago_historical}, have been grouped according to their visual similarities:

\begin{enumerate}[label=(\alph*)]
    \item \textit{Work}: photos belonging to this class are mostly characterized by people sitting and/or standing in workplaces and wearing work clothes;
    \item \textit{Free-time}, which includes the \textit{Free-time \& Holidays} categories: such category investigates the forms and ways of experiencing leisure time, reconstructing, wherever possible, generational and gender differences. It also includes images representing people which make new experiences, visiting far off landmarks, expanding social relationships and interacting with nature;
    \item \textit{Motorization}: although often closely related to the \textit{Free-time} category, this class has been distinguished as it includes symbolic objects such as cars and motorcycles, which represent a social and historical landmark;
    \item \textit{Music}: as for the \textit{Motorization} one, this class may also include scenes from leisure time, characterized in this case by the appearance of musical instruments or events;
    \item \textit{Fashion}, which includes the \textit{Fashion \& Traditions} categories: in this, clothing represents a mirror of the articulated intertwining of socio-economic, political and cultural phenomena. This class is characterized by the presence of symbolic objects and clothes, such as suits, trousers, skirts and coats;
    \item \textit{Affectivity}, which includes the \textit{Affectivity \& Friendship \& Family} categories: such photographs are characterized by the presence of people (e.g., couples, friends, families or colleagues) bound by inter-personal relationships;
    \item \textit{Rites}, which includes the \textit{Rites \& Marriages} categories: in the pictures comprised in this category, it is possible to find portraits pertaining the sacred and/or celebratory events which characterize the life of a family;
    \item \textit{School}: this class includes all the photos which represent schools, often characterized by symbolic objects (e.g., desk, blackboard) or group of students;
    \item \textit{Politics}: this class contains photos related to political gatherings, demonstrations and events.
\end{enumerate}

Fig.~\ref{fig:distr_imago_normal} reports the number of labelled images available per year in the 1930 to 1999 time frame (out of such time interval, the number of available images is too little to be represented visually). Such Figure also exhibits the unbalance observed both in terms of number of photos per year (the greatest share is concentrated between 1950 and 1980) and in terms of their socio-historical classification, as the dominant classes result to be the \textit{Affectivity}, \textit{Fashion} and \textit{Free-time} ones. Fig.~\ref{fig:samples_imago} shows four exemplar images from the IMAGO dataset which belong to different decades and represent different socio-historical contexts. 
From this Figure, it is possible to appreciate the different characteristics that can be found in each photo (e.g., number of people, clothing, colors and location).

The following Section reports upon the deep leaning models exploited in our analysis.

\section{Deep Learning models}
\label{sec:model}
In this Section we firstly describe the image pre-processing actions carried out on the IMAGO dataset. Secondly, we detail how the training, validation and test sets have been composed to perform a fair evaluation of the models. Finally, we describe the considered deep neural network models along with their architecture, the training settings and the evaluation protocol.

\subsection{Pre-processing}
\label{subsec:pre_processing}
The pre-processing phase aimed at: (i) isolating the regions of interest which could enhance the performance of the selected deep learning models, (ii) improving the quality of the images composing the dataset, resorting to different computer vision algorithms.

\begin{figure*}[t]
   \begin{subfigure}{0.45\linewidth}
     \includegraphics[width=\linewidth]{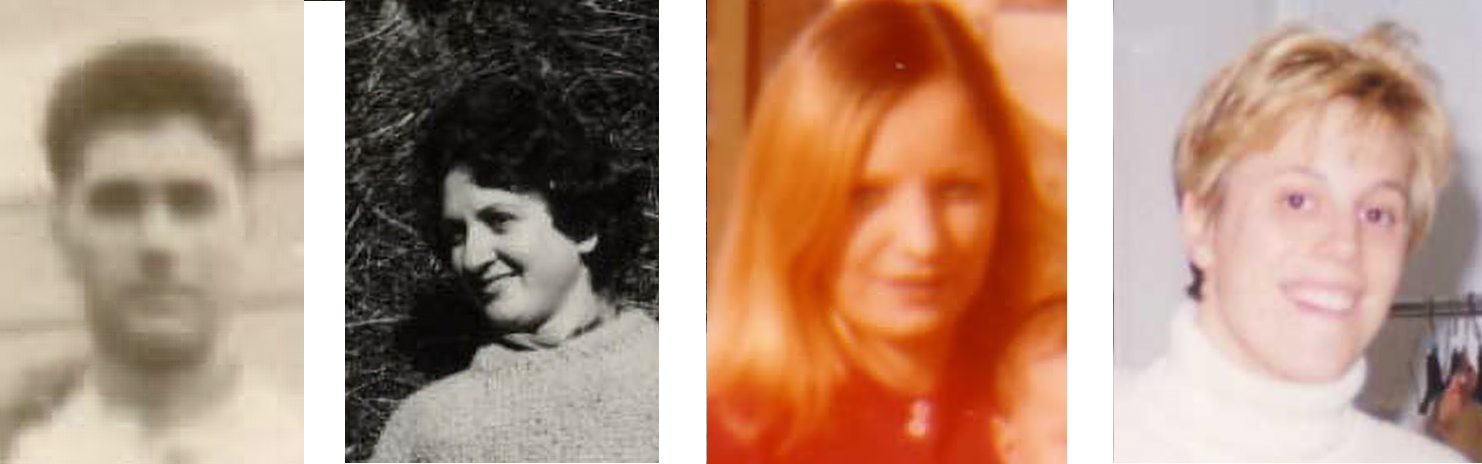}
   \caption{IMAGO-FACES}
     \label{fig:sample_imago_face}
   \end{subfigure}
   \hfill
   \begin{subfigure}{0.5\linewidth}
     \includegraphics[width=\linewidth]{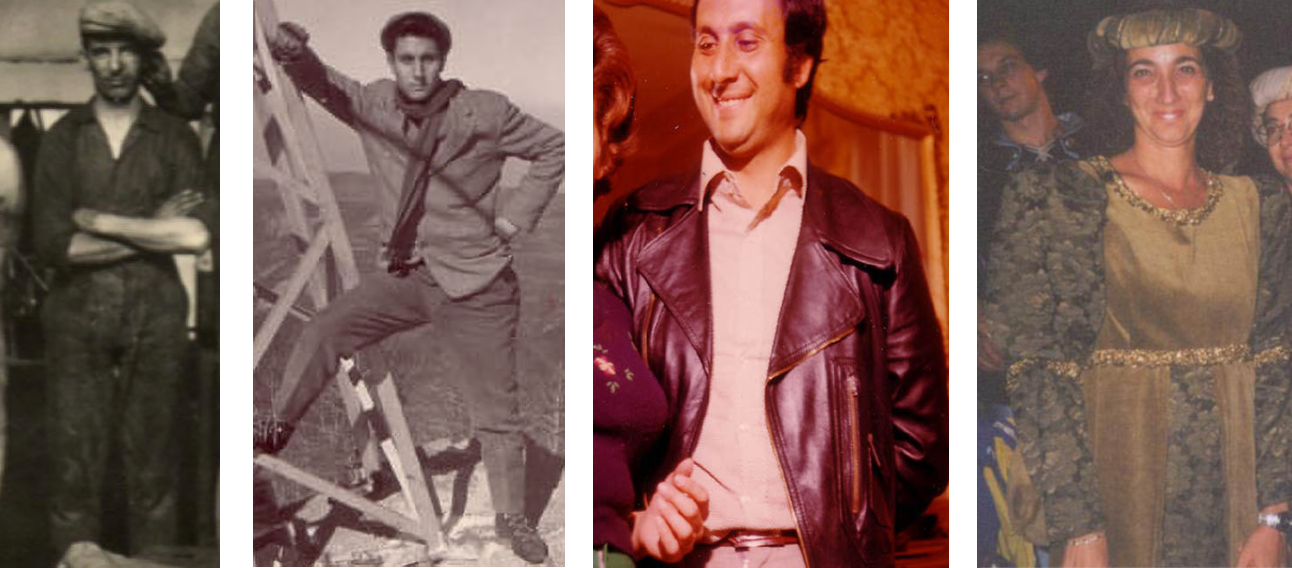}
   \caption{IMAGO-PEOPLE}
     \label{fig:sample_imago_people}
   \end{subfigure}
   \caption{IMAGO-FACES and IMAGO-PEOPLE samples.}
\end{figure*}

As reported in Section~\ref{sec:related-work}, both faces and people represent regions of interest to be exploited for the dating analysis~\cite{salem2016analyzing,ginosar2015century}.
Following such insight, we created the IMAGO-FACES and the IMAGO-PEOPLE datasets, comprising each over 60,000 samples: the first composed of individual faces, the second of single person's full figure images. In particular, we processed each image of the IMAGO dataset using an open-source implementation of YOLO-FACE~\cite{yolo_face} and YOLO~\cite{yolo}, respectively.

The IMAGO-FACES dataset has been constructed accounting for the number of people portrayed in a photo. In fact, adopting a fixed size bounding box it may be possible to loose relevant details (e.g., hairstyle) or to include pixels related to the faces of other people. To avoid such problem, an adaptive strategy has been adopted: the size of the bounding box used to crop a face depends on the number of people portrayed in a photo, the greater the number of people, the smaller the bounding box. In this way, it was possible to extract the shoulders and the full head of a single person even when a picture portrayed tens of people. Fig.~\ref{fig:sample_imago_face} shows some sample images taken from the IMAGO-FACES dataset considering different decades and different socio-historical contexts. The construction of the IMAGO-PEOPLE dataset follows the same criteria employed for IMAGO-FACES, though, images can present different aspect ratios (i.e., people may be standing or sitting in photos). Fig.~\ref{fig:sample_imago_people} shows exemplar images from IMAGO-PEOPLE. It is possible to appreciate that IMAGO-PEOPLE includes details that are not present in IMAGO-FACES (e.g., the clothing of a person).

\begin{figure}
    \centering
    \includegraphics[width=\linewidth]{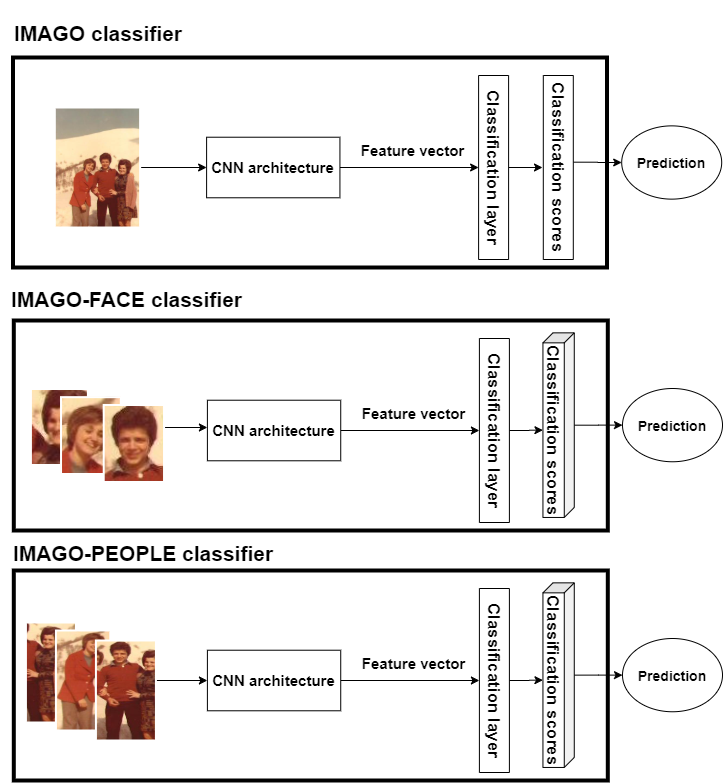}
    \caption{Model architectures for single-input classifiers.}
    \label{fig:model_single_image}
\end{figure}

Finally, we verified the utility of performing denoising and super resolution operations, as all the images considered in this work derive from scans of analog prints. For denoising we tested the neural network model from~\cite{zhang2018ffdnet} and the Bilateral Filter~\cite{bilateral_filtering}. For super resolution, we used an open-source implementation of the ESRGAN model~\cite{wang2018esrgan} within the Image Restoration Toolbox~\cite{kair_toolbox}. The overall improvement obtained adopting such strategies revealed to be negligible, we hence opted for an analysis based on the original scans of analog photos.

\subsection{Generation of train, validation and test subsets}
\label{subsec:train_validation_test}
 
All relevant datasets (IMAGO, IMAGO-FACES and IMAGO-PEOPLE) have been partitioned in three subsets or groups of pictures: 80\% for training and 20\% for testing. In addition, 10\% of the training images is parted to be used as the validation subset for the tuning of model hyperparameters. In particular, for each image in the train set of IMAGO, the faces and the people there portrayed are extracted and added to the corresponding IMAGO-FACES and IMAGO-PEOPLE subsets, respectively. This process is repeated also for the validation and test sets, as it guarantees that none of the training samples may end in the validation and test sets.

\begin{figure*}
    \centering
     \includegraphics[width=0.86\linewidth]{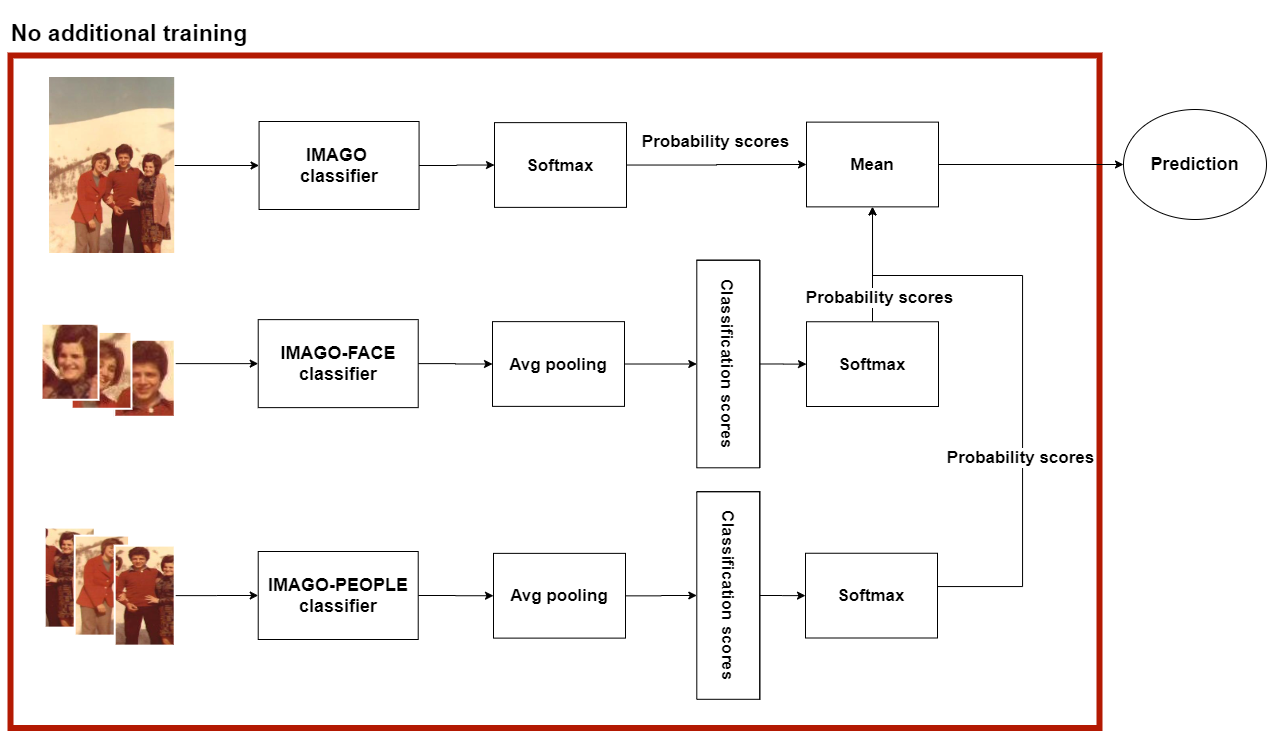}
     \caption{Dating task: Ensemble classifier model architecture.}
    \label{fig:combined_model_ensemble}
\end{figure*}

\begin{figure*}
    \centering
     \includegraphics[width=0.86\linewidth]{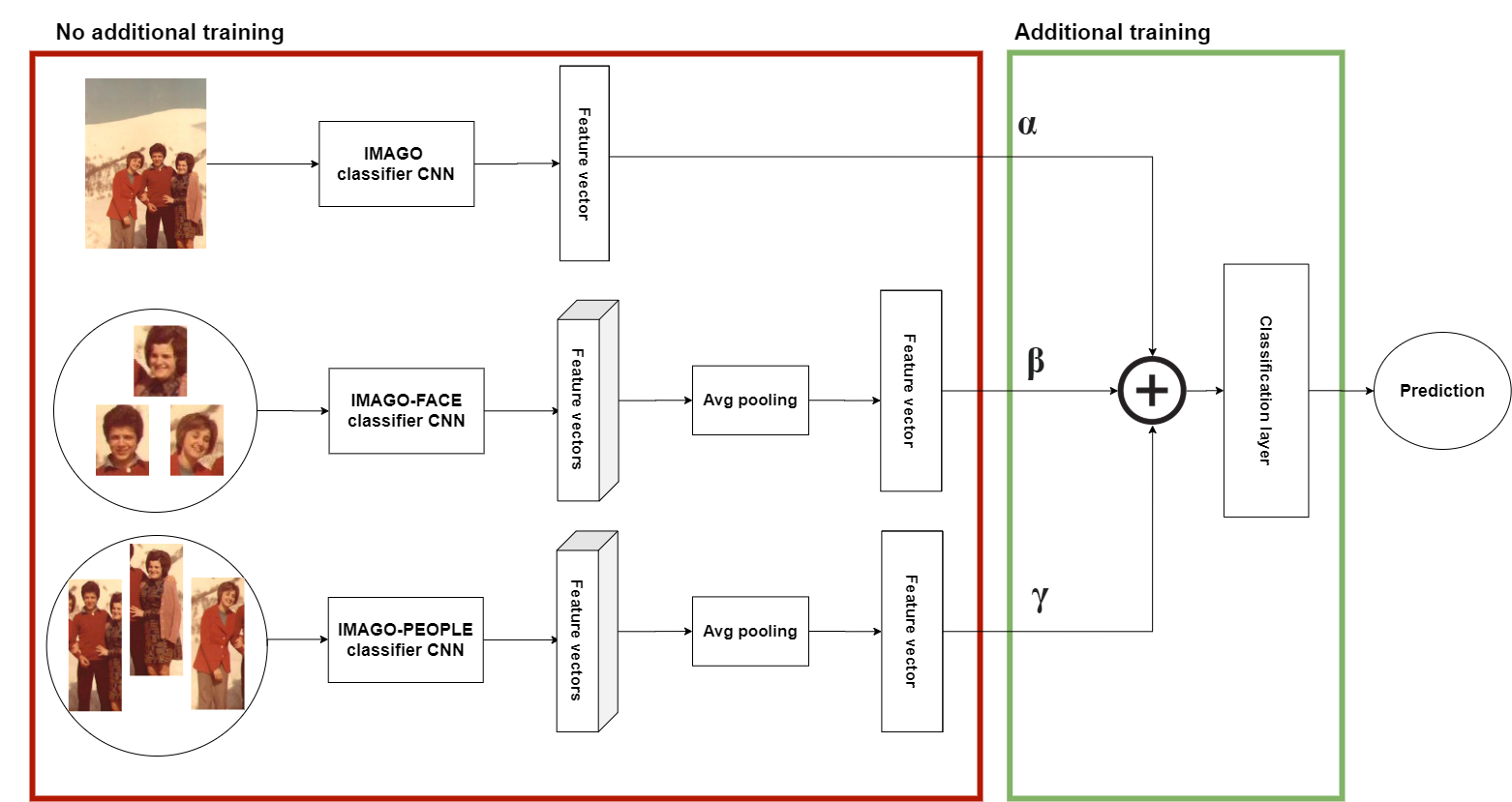}
     \caption{Dating task: Merged classifier model architecture.}
    \label{fig:combined_model_merged}
\end{figure*}

\subsection{Model architectures}
\label{subsec:model_architectures}

In this work we considered single and multi-input deep learning architectures for both the dating and socio-historical context classification tasks. The former type of models analyzes the images pertaining one of the datasets of interest in isolation, the latter, instead, operates on their combination (i.e., images in IMAGO, IMAGO-FACES and IMAGO-PEOPLE). 

The deep neural network models exploited in our experiments are three well-known CNN architectures pre-trained on ImageNet~\cite{imagenet}: (i) Resnet50~\cite{resnet}, (ii) InceptionV3~\cite{inceptionv3} and (iii) DenseNet121~\cite{densenet121}. Each of such models was modified replacing the top-level classifier with a new classification layer, whose structure depends on the task (i.e., the number of output classes) and whose weights have been randomly initialized. 
The pre-trained convolutional layers have been specifically fine-tuned for the given input data and task (dating or socio-historical context classification).

For what regards the single-input classifiers, one has been trained per each dataset. In particular, Fig.~\ref{fig:model_single_image} exhibits the architecture laid for IMAGO, IMAGO-FACES and IMAGO-PEOPLE.

The multi-input deep learning architectures result from the combination of the single-input models: the aim is to verify whether the classification performance improves with the combination of their knowledge.
We resorted at first to an Ensemble model which puts together the single-input classifiers introduced before, without performing any additional training. Adopting such architecture, shown in Fig.~\ref{fig:combined_model_ensemble}, the score assigned to each image belonging to IMAGO, and to its corresponding faces (in IMAGO-FACES) and full figures (in IMAGO-PEOPLE), is stored and subsequently combined to improve the class prediction. In fact, as a picture could contain more than one person, multiple IMAGO-FACES and IMAGO-PEOPLE images could stem from a single IMAGO sample. 
Thus, each image in IMAGE-FACES and IMAGO-PEOPLE, produces a classification score vector by means of their respective single-input classifiers (one per each branch). The score vectors obtained are averaged out and normalized. The three vectors, resulting from the three models, are finally averaged out providing the output from which the most probable class is inferred.

We then defined the multi-input Merged model shown in Fig.~\ref{fig:combined_model_merged}, which with respect to the Ensemble model aims at not only putting together different sources of information, but also at learning how. Hence, a new training session is carried out as the newly introduced network is asked to learn how to perform such combination. 
As before, the already trained single-input classifiers are employed, one along each branch (IMAGO, IMAGE-FACES, IMAGO-PEOPLE), but the classification layer is removed, preserving the CNN backbone. 
The cardinality of the different features depends on the number of faces/people (in IMAGE-FACES, IMAGO-PEOPLE) portrayed in an image. For this reason the average of such feature vectors is computed, to be able to combine them with the vector obtained from the original image (in IMAGO). The three resulting feature vectors, one for the original image, one for faces and one for people, are then linearly combined by means of a weighted sum, whose weights are the real scalars $\alpha$, $\beta$ and $\gamma$, whose values are learned during the training phase.
The feature vector resulting from the linear combination is then fed to a fully connected layer with a softmax activation yielding the final probability vector and, consequently, the most probable class. Summing up, this model differs from the Ensemble classifier as it extracts the feature vectors instead of the scores from the pre-trained single-input classifiers and uses trainable weights to balance their importance. Moreover, it contains an additional dense classification layer which is trained from scratch observing the whole training set.

\subsection{Training settings}
During training we applied random cropping and horizontal flipping in order to make the model less prone to overfitting. Each model has been fine-tuned using a weighted cross entropy loss and an Adam optimizer with a learning rate of 1e-4 and a weight decay of 5e-4.
We set the batch size to 32 for the training of the IMAGO classifier and to 64 for the training of the IMAGO-FACES and IMAGO-PEOPLE models.

\subsection{Model comparison}
To compare the performance of single and multi-input classifiers, we need to establish an evaluation metric for those photos that contains more than one person. 

Simply, for IMAGO-FACES and IMAGO-PEOPLE classifiers, we evaluate the accuracy not for a single face or person, but agglomerating votes for each of these in one picture. This means that, if a picture contains $n$ persons, the  IMAGO-FACES (and IMAGO-PEOPLE) classifier prediction would be considered correct if the unique vote that derive from the combination of the $n$ extracted is correct. Practically, we average the probability vectors returned by the single-input classifier for each image and then we compute the most probable class.

\section{Experimental Results}
\label{sec:results}
This Section reports on the results obtained for both the dating and the socio-historical context classification tasks described so far. 
For what concerns image dating, 15,673 pictures, covering the 1930 to 1999 temporal interval, have been employed to reduce the imbalance problem of the IMAGO dataset that is even more challenging than the one exhibited in Fig. \ref{fig:distr_imago_normal}. The entire dataset (16,642 photos spanning over the 1845-2009 time period) is instead used during the analysis of the socio-historical context task.

\subsection{Dating task results}
\label{sec:results_dating}
The evaluation of the performance of the dating model is computed exploiting time distances, as also reported in~\cite{salem2016analyzing,ginosar2015century}. The time distance defines the tolerance accepted in predictions with respect to the actual year. For example, if a photo was labeled with year 1932 and the model returned 1927 (or even 1937), this would be considered correct for those cases where the time distance was set to values equal or greater than 5 and wrong otherwise.
In this work, model accuracies were computed considering temporal distances of 0, 5 and 10 years.

The results are reported in Table~\ref{tab:results_classifiers_dating}.
It is possible to appreciate that different baseline models return similar accuracy values. In addition, the same baselines returned comparable results when exploiting different datasets (i.e., IMAGO, IMAGO-FACES and IMAGO-PEOPLE). This may depend on many factors, including the people's appearance (e.g., dresses, hairstyle, earrings, trousers), as discussed in Section \ref{sec:discussion}. When comparing the results of the different approaches, multi-input improves with respect to single-input and the Merged one emerges above all. Evidently, the complementary knowledge exploited with multi-input models contributes to such results. This is found observing both the negative trend of the mean errors and the positive one of accuracies. 

\begin{table*}[h] \large
\centering
\renewcommand{\arraystretch}{2}
\begin{adjustbox}{width=\linewidth}
     \begin{tabular}{cc|ccc|ccc|ccc|ccc|ccc|}
        \multicolumn{2}{c}{} & \multicolumn{15}{c}{\textbf{Classifier}} \\
        \Cline{2pt}{3-17}
        \multicolumn{2}{c}{} & \multicolumn{3}{|c}{\textbf{Image}} & \multicolumn{3}{|c}{\textbf{Face}} & \multicolumn{3}{|c}{\textbf{People}} & \multicolumn{3}{|c}{\textbf{Ensemble}} & \multicolumn{3}{|c|}{\textbf{Merged}} \\
        \Cline{2pt}{3-17}
        \multicolumn{2}{c|}{} & \textbf{ResNet50} & \textbf{InceptionV3} & \textbf{DenseNet121} & \textbf{ResNet50} & \textbf{InceptionV3} & \textbf{DenseNet121} & \textbf{ResNet50} & \textbf{InceptionV3} & \textbf{DenseNet121} & \textbf{ResNet50} & \textbf{InceptionV3} & \textbf{DenseNet121} & \textbf{ResNet50} & \textbf{InceptionV3} & \textbf{DenseNet121} \\
        \specialrule{2pt}{1pt}{1pt}
        \multirow[c]{3}{*}{\textbf{Accuracy}} &
        \textbf{d = 0} &
        \textbf{11.31} & 10.45 & 10.68 &
        \textbf{15.01} & 14.60 & 12.91 &
        \textbf{15.77} & 12.56 & 13.99 &
        \textbf{17.78} & 13.58 & 15.68 &
        \textbf{18.71} & 17.14 & 16.22 \\
        \cline{2-17}
        & \textbf{d = 5} &
        \textbf{62.56} & 61.38 & 60.77 &
        \textbf{58.09 }& 56.95 & 57.81 &
        \textbf{62.40} & 60.04 & 59.69 &
        \textbf{63.99} & 59.94 & 63.00 &
        \textbf{67.59} & 67.56 & 66.67 \\
        \cline{2-17}
        & \textbf{d = 10} &
        82.54 & \textbf{82.82} & 82.47 &
        78.39 & 78.46 & \textbf{79.70} &
       \textbf{ 82.47} & 81.39 & 81.42 &
        \textbf{83.33}& 80.69 & 82.98 &
        86.17 & \textbf{86.30} & 86.07 \\
        \specialrule{2pt}{1pt}{1pt}
        \multirow[c]{1}{*}{\vtop{\hbox{\strut \textbf{Error}}\hbox{\strut \textbf{(d = 0)}}}} &
        \textbf{Mean} &
        \textbf{6.01} $\pm$ \textbf{6.76}& 6.20 $\pm$ 6.89 & 6.13 $\pm$ 6.74 &
        6.56 $\pm$ 7.13 & 7.08 $\pm$ 7.20 & \textbf{6.54 $\pm$ 6.92} &
        6.82 $\pm$ 5.98 & \textbf{6.20 $\pm$ 6.75} & 6.23 $\pm$ 6.82 &
        \textbf{5.63 $\pm$ 6.51} & 6.34 $\pm$ 7.08 & 5.86 $\pm$ 6.68 &
        \textbf{5.06 $\pm$ 6.01} & 5.14 $\pm$ 5.92 & 5.22 $\pm$ 5.95 \\
        \specialrule{2pt}{1pt}{1pt}
    \end{tabular}
\end{adjustbox}
\caption{Accuracy: dating models accuracies for different time distances (d = 0, d = 5, d = 10). Error (d = 0): Mean errors computed for a time distance equal to 0.}
\label{tab:results_classifiers_dating}
\end{table*}

In the analyses that follow, the Merged model was selected utilizing as backbone reference the ResNet50. Such choices were made because the Merged model provided the best results and the ResNet50 the best trade-off between accuracy and model-dimensionality ~\cite{coleman2019analysis}.

\begin{figure}[h!]
    \centering
    \includegraphics[width=0.85\linewidth]{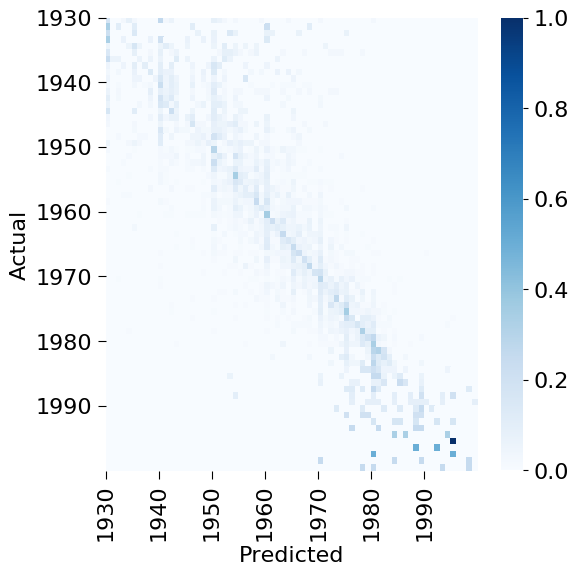}
    \caption{Confusion matrix for the dating task obtained using the Merged classifier considering a time distance equal to 0.}
    \label{fig:merged_classifier_dating}
\end{figure}

Fig.~\ref{fig:merged_classifier_dating} shows the confusion matrix considering a time distance equal to 0.
The diagonal structure demonstrates that the confusion mostly occurs between neighboring years, except for the initial and the final decades, as also highlighted in other works~\cite{ginosar2015century}.
The confusion created within the first 20 years could be caused by the low quality of the images and the limited number of samples representing those years.
This phenomenon could also be explained considering two additional facts: (i) during those decades photos mainly fall in two classes (i.e., \textit{Fashion} and \textit{Affectivity}, as clear from Figure \ref{fig:distr_imago_normal}), and (ii) the percentage of photos shot in a professional setting was much higher than in the decades that followed. The confusion created within the last 20 years, instead, could be related to the fact that the number of images for these years is very limited (as shown in Fig.~\ref{fig:distr_imago_normal}).

\begin{figure}[h]
    \centering
    \includegraphics[width=\linewidth]{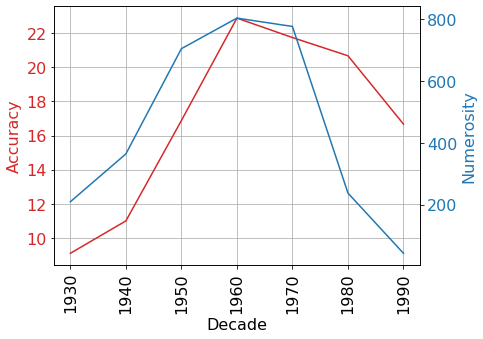}
    \caption{Merged model accuracy (red line) and number of samples (blue line) by decade for a time distance equal to 0.}
    \label{fig:merged_classifier_dating_decades}
\end{figure}

Nevertheless, it is interesting to observe the information provided in Fig.~\ref{fig:merged_classifier_dating_decades}, where the model accuracy and the number of samples per decade are reported. This Figure confirms the finding exhibited by the confusion matrix, the model accuracy improves after the 50's. 
Fig.~\ref{fig:merged_classifier_dating_decades} also shows that, despite a reduction in terms of available samples per decade after the 80's, the performance of the model has a slight decrease.
In fact, it is evident that accuracy generally improves after the 50's (also when the number of samples drops), again, this could be related to the fact that the images are of better quality with respect to the previous decades.

Another behavior that was observed is that the accuracy of the model grows as the number of faces and/or people increases within a photo. To further verify such phenomenon, we decided to use the Ensemble classifier (described in Fig.~\ref{fig:combined_model_ensemble}), as this model let us isolate the contribution of each single input branch (IMAGO, IMAGO-FACES, IMAGO-PEOPLE). To this aim, an experiment was conducted considering a test subset characterized by a fixed and constant number $n$ of faces/people across all images. To weigh the role of the number of faces/people, fixing $n = 8$, the accuracy values were computed considering $k$ faces/people, with $k$ growing from $1$ to $n$. In essence, for all those images which comprised 8 faces/people, we computed the dating accuracy that could be obtained as a function of the number of faces/people considered for its evaluation. 
To ensure the completeness and fairness of this experiment, all possible combinations of the $k$ faces/people were considered. Accuracy results have been grouped by $k$ and are reported in Table~\ref{tab:increasing_face_people_accuracies_dating_classifier}. In order to verify that the increasing dating accuracy was solely due to an increase of the number of considered faces/people, we proceeded to isolate the influence of the original image (IMAGO dataset).
To this aim, the date of an image was evaluated, including and excluding the contribution of the IMAGO classifier in Fig.~ \ref{fig:combined_model_ensemble}: (i) verifying the contributions of the IMAGO-FACES and IMAGO-PEOPLE classifiers in isolation and (ii) verifying the contributions of the IMAGO-FACES and IMAGO-PEOPLE classifiers while both active.
While confirming that a higher number of faces or people can improve the overall accuracy, it is also possible to see that the contribution of the original image decreases with a higher number of faces and people. It is worth highlighting that this experiment has been conducted on a subset of the test set.

\begin{table*}[h] \large
\centering
\renewcommand{\arraystretch}{2}
\begin{adjustbox}{width=0.8\linewidth}
     \begin{tabular}{cc|c|c|c|c|c|c|}
        \multicolumn{2}{c}{} & \multicolumn{6}{c}{\textbf{Ensemble classifier with ResNet50 backbone}} \\
        \Cline{2pt}{3-8}
        \multicolumn{2}{c|}{} &  \multicolumn{2}{c|}{\textbf{IMAGO-FACE}} &  \multicolumn{2}{c|}{\textbf{IMAGO-PEOPLE}} &  \multicolumn{2}{c|}{\textbf{IMAGO-FACE w/ IMAGO-PEOPLE}} \\
        \Cline{2pt}{3-8}
        \multicolumn{1}{c}{} & \textbf{Number of faces/people} & \textbf{w/o IMAGO} & \textbf{w/ IMAGO} & \textbf{w/o IMAGO} & \textbf{w/ IMAGO} & \textbf{w/o IMAGO} & \textbf{w/ IMAGO} \\
        \specialrule{2pt}{1pt}{1pt}
        \multirow[c]{8}{*}{\textbf{Accuracy}} & 1 &  15.25 & 16.75 & 13.99 & 15.03 & 24.55 & 24.78 \\
        \cline{2-8}
         & 2 & 15.57 & 17.21 & 18.03 & 19.05 & 26.62 & 27.50 \\
        \cline{2-8}
         & 3 & 17.04 & 18.50 & 18.77 & 19.49 & 28.06 & 28.79 \\
        \cline{2-8}
         & 4 & 17.54 & 19.31 & 19.49 & 19.54 & 29.60 & 30.12 \\
        \cline{2-8}
         & 5 & 17.61 & 19.57 & 19.96 & 19.45 & 30.68 & 31.02 \\
        \cline{2-8}
         & 6 & 17.93 & 20.00 & 20.24 & 19.43 & 31.39 & 31.60 \\
         \cline{2-8}
         & 7 & 17.75 & 20.25 & 20.39 & 20.09 & 31.92 & 31.99 \\
         \cline{2-8}
         & 8 & 18.00 & 20.00 & 20.24 & 21.43 & 33.33 & 33.33 \\
        \specialrule{2pt}{1pt}{1pt}
    \end{tabular}
\end{adjustbox}
\caption{Accuracy results for increasing number of faces/people exploiting the Ensemble classifier with ResNet50 backbone for a time distance equal to 0.}
\label{tab:increasing_face_people_accuracies_dating_classifier}
\end{table*}

\subsection{Socio-historical context classification task results}
\label{subsec:result_socio_historical}
In this Section we analyze the results that were obtained for the socio-historical context classification task using state of the art CNN architectures.
The results are reported in Table~\ref{tab:results_classifiers_socio_cultural}, in terms of top-$k$ accuracy: if the correct class is not the one with the highest predicted probability, but falls among the $k$ with the highest predicted probabilities, it will be counted as correct. It is interesting to compare the performance of the single-input classifiers in such terms: the IMAGO classifier exhibits a top-3 accuracy of 92.85\% with a ResNet50 backbone, while, in the same setting, the other single-input classifiers obtain a lower performance. 
The implications of such result on the socio-historical analysis is discussed in Section~\ref{subsec:socio_gradcam}.

\begin{table*}[h] \large
\centering
\renewcommand{\arraystretch}{1.8}
\begin{adjustbox}{width=0.8\linewidth}
     \begin{tabular}{c|ccc|ccc|ccc|}
        \multicolumn{1}{c}{} & \multicolumn{9}{c}{\textbf{Classifier}} \\
        \Cline{2pt}{2-10}
        \multicolumn{1}{c}{} & \multicolumn{3}{|c}{\textbf{IMAGO}} & \multicolumn{3}{|c}{\textbf{IMAGO-FACE}} & \multicolumn{3}{|c|}{\textbf{IMAGO-PEOPLE}} \\
        \Cline{2pt}{2-10} 
        \multicolumn{1}{c|}{} & \textbf{ResNet50} & \textbf{InceptionV3} & \textbf{DenseNet121} & \textbf{ResNet50} & \textbf{InceptionV3} & \textbf{DenseNet121} & \textbf{ResNet50} & \textbf{InceptionV3} & \textbf{DenseNet121} \\
        \specialrule{2pt}{1pt}{1pt}
        \textbf{Top-1} &
        \textbf{64.35} & 64.08 & 63.72 &
        41.30 & \textbf{43.37} & 41.39 &
        56.54 & \textbf{57.20} & 55.64 \\
        \hline
        \textbf{Top-2} &
        \textbf{85.00} & 83.83  & 83.38 &
        65.55 & \textbf{65.92} & 65.28 &
        78.48 & 77.34 & \textbf{78.90} \\
        \hline
        \textbf{Top-3} &
        \textbf{92.85} & 92.28 & 92.37 &
        \textbf{82.75 }& 81.18 & 82.66 &
        \textbf{89.90} & 89.36 & 89.87 \\
        \hline
        \textbf{Top-4} &
        96.66 & \textbf{96.75} & 96.54 &
        90.86 & 90.20 & \textbf{91.28} &
        \textbf{94.74} & 94.47 & 94.53 \\
        \hline
        \textbf{Top-5} &
        98.35 & \textbf{98.53} & 98.47 &
        94.98 & 95.01 & \textbf{95.31} &
        \textbf{97.42} & 97.14 & 97.26 \\
        \specialrule{2pt}{1pt}{1pt}
    \end{tabular}
\end{adjustbox}
\caption{Socio-historical models accuracies for an increasing Top-$k$ classification ($k$ ranging from 1 to 5).}
\label{tab:results_classifiers_socio_cultural}
\end{table*}

Please note that we here concentrated on the use of single-input architectures analyzing the IMAGO dataset, as multi-input ones did not provide any significant contribution to the fulfillment of the task at hand. Why this occurs may be explained as follows: the socio-historical context represented by an image may be inferred with a greater precision when it is analyzed as a whole, including its spatial relationships, objects and locations, in addition to how people look. In other words, the performance of the IMAGO classifier naturally emerges with respect to the results obtained utilizing the other single-input ones (IMAGO-FACES, IMAGO-PEOPLE), as shown in Table~\ref{tab:results_classifiers_socio_cultural}. In essence, the spatial relationships that are broken when constituting the IMAGO-FACES and IMAGO-PEOPLE datasets are key for this task. This is also why multi-input classifiers may not perform as well as the IMAGO classifier.
Reminding that the final output of multi-input classifiers is a combination of the knowledge acquired along the three branches, we can observe that the information acquired with the IMAGO classifier may be corrupted by the information extracted from the other single-input classifiers.

\begin{figure}[h]
    \centering
    \includegraphics[width=\linewidth]{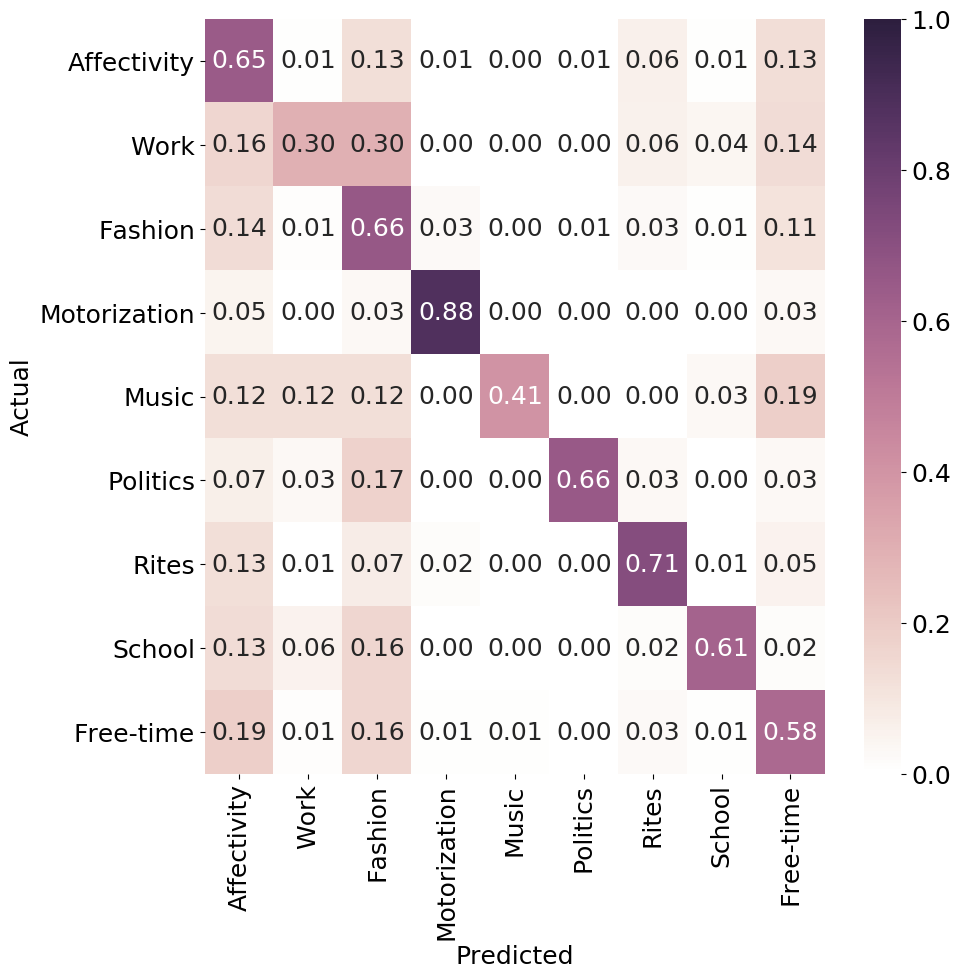}
    \caption{Confusion matrix for total image classifier.}
    \label{fig:conf_matrix_semantic}
\end{figure}

Fig.~\ref{fig:conf_matrix_semantic} shows the confusion matrix obtained with the IMAGO classifier. It is possible to observe that three classes are the responsible for the largest share of misclassifications, in fact, the model is often misled to choose one among \textit{Fashion}, \textit{Affectivity} and \textit{Free-time}, instead of the correct class.
This can be justified considering that all the other classes contain images which are not easily separable from the three aforementioned ones. Indeed, pictures belonging to other classes can be visually similar (e.g., a model at work and/or people embracing during a concert, could be classified as \textit{Fashion} and \textit{Affectivity} while belonging to the \textit{Work} and \textit{Music} classes, respectively) and hence difficult to discriminate.
This is also true for the \textit{Free-time} class, which can be considered the most general one because photos that fall within this category are characterized by people that are involved in different kind of activities in heterogeneous environments.
Another interesting fact is that the \textit{Work} class is mostly confused with the \textit{Fashion} and \textit{Affectivity} ones. This could be explained since the images that fall within the \textit{Work} class represent groups of people wearing work-suits and/or being in working environments (e.g., people in a cafeteria, market or factory).
A similar phenomenon takes place for the \textit{Music} class. In particular, the performance of the model related to this class is slightly better than the one exhibited for the \textit{Work} one. This may be traced back to the confusion existing among the \textit{Free-Time}, \textit{Work}, \textit{Fashion} and \textit{Affectivity} classes. This confusion was generated by the fact that many photos belonging to the \textit{Music} class included different aspects of daily life: (i) a band playing at a concert which may be classified as \textit{Music} or \textit{Work}, or, (ii) a group of friends at a concert which could belong to \textit{Music} or \textit{Free-Time} depending on the perspective. This will be further analyzed in Section \ref{subsec:socio_gradcam}.

\section{What do the models learn?}
\label{sec:discussion}

The aim now is to understand which cues led the trained models to determine the date of a picture or to individuate its socio-historical context. To do so, in the following the Grad-Cam algorithm~\cite{grad_cam} is applied: in brief, this algorithm delimits the areas of an image exploited by deep learning models for the classification.

\subsection{Dating task}

\begin{figure*}[h]
\centering
\begin{adjustbox}{width=0.85\linewidth}
\begin{tabular}{m{12mm}!{\vrule width 2pt}m{24mm}!{\vrule width 2pt}m{24mm}m{24mm}!{\vrule width 2pt}m{24mm}m{24mm}!{\vrule width 2pt}}
    \multicolumn{1}{c!{\vrule width 2pt}}{} & \multicolumn{1}{c!{\vrule width 2pt}}{\textbf{IMAGO}} & \multicolumn{2}{c!{\vrule width 2pt}}{\textbf{IMAGO-FACES}} & \multicolumn{2}{c!{\vrule width 2pt}}{\textbf{IMAGO-PEOPLE}} \\
    \textbf{1930} & 
    \includegraphics[width=24mm,height=24mm]{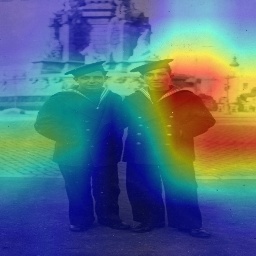}
    & \includegraphics[width=24mm,height=24mm]{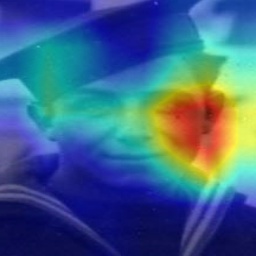} &   \includegraphics[width=24mm,height=24mm]{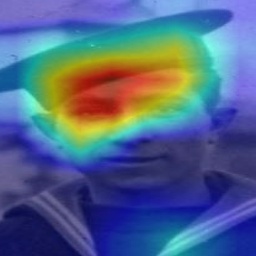} &   \includegraphics[width=24mm,height=24mm]{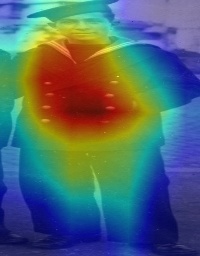} &   \includegraphics[width=24mm,height=24mm]{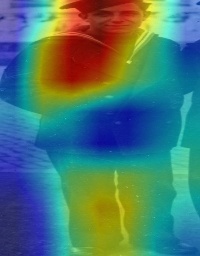} \\
    \textbf{1940} & 
    \includegraphics[width=24mm,height=24mm]{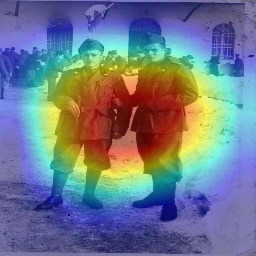}
    & \includegraphics[width=24mm,height=24mm]{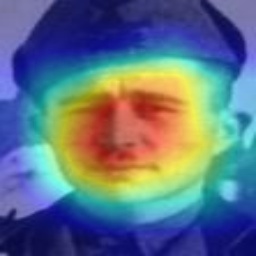} &   \includegraphics[width=24mm,height=24mm]{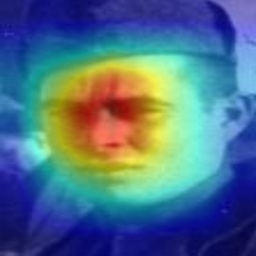} &   \includegraphics[width=24mm,height=24mm]{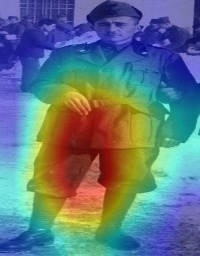} &   \includegraphics[width=24mm,height=24mm]{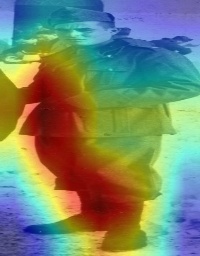} \\
    \textbf{1950} & \includegraphics[width=24mm,height=24mm]{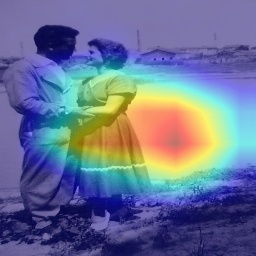} &   \includegraphics[width=24mm,height=24mm]{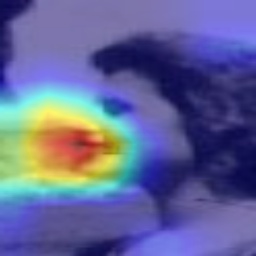} &   \includegraphics[width=24mm,height=24mm]{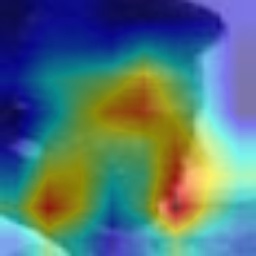}&   \includegraphics[width=24mm,height=24mm]{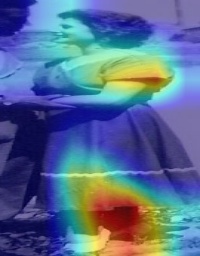}&   \includegraphics[width=24mm,height=24mm]{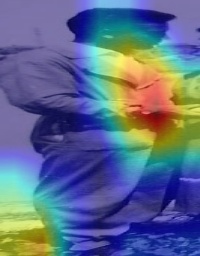} \\
    \textbf{1960} &  \includegraphics[width=24mm,height=24mm]{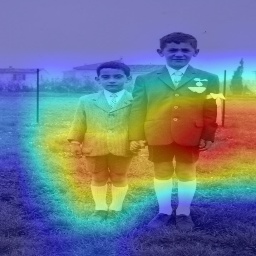} &   \includegraphics[width=24mm,height=24mm]{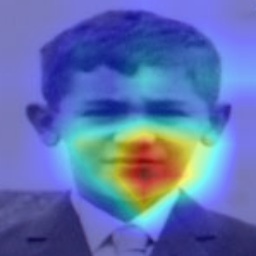} &   \includegraphics[width=24mm,height=24mm]{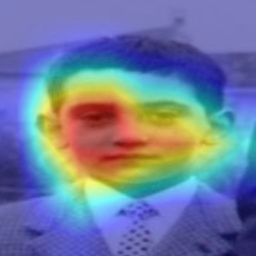} &   \includegraphics[width=24mm,height=24mm]{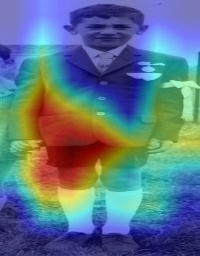} &   \includegraphics[width=24mm,height=24mm]{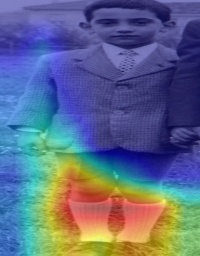} \\
    \textbf{1970} & \includegraphics[width=24mm,height=24mm]{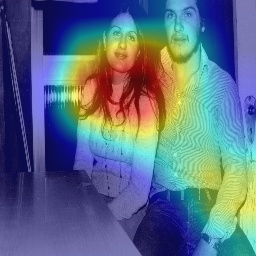} &   \includegraphics[width=24mm,height=24mm]{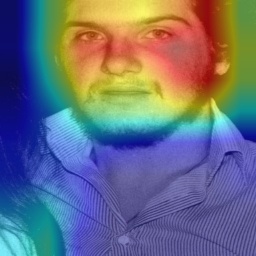} &   \includegraphics[width=24mm,height=24mm]{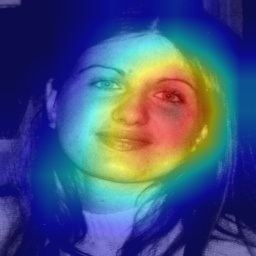} & 
    \includegraphics[width=24mm,height=24mm]{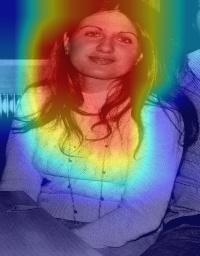} & 
    \includegraphics[width=24mm,height=24mm]{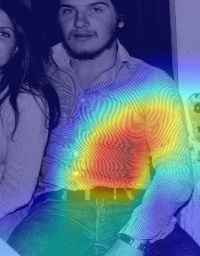} \\
    \textbf{1980} & \includegraphics[width=24mm,height=24mm]{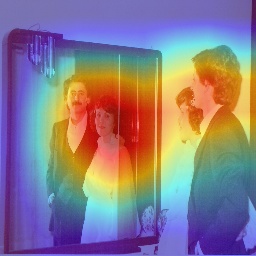} &   \includegraphics[width=24mm,height=24mm]{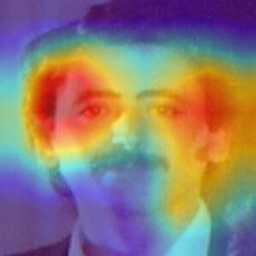} &   \includegraphics[width=24mm,height=24mm]{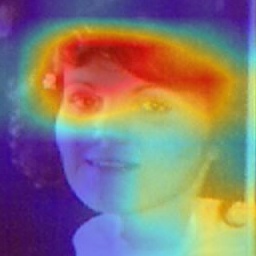} &
    \includegraphics[width=24mm,height=24mm]{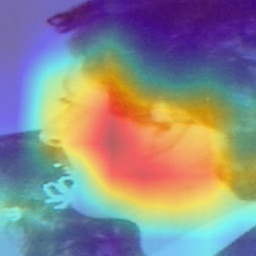} &  \includegraphics[width=24mm,height=24mm]{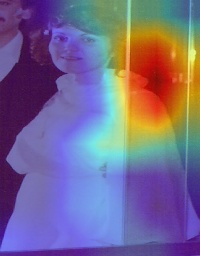} \\
    \textbf{1990} & \includegraphics[width=24mm,height=24mm]{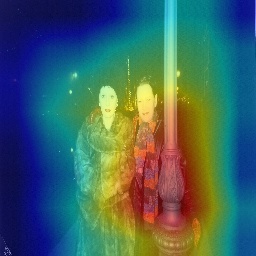} &   \includegraphics[width=24mm,height=24mm]{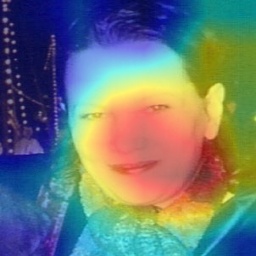} &   \includegraphics[width=24mm,height=24mm]{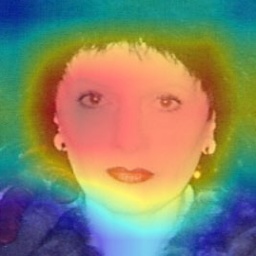} &
    \includegraphics[width=24mm,height=24mm]{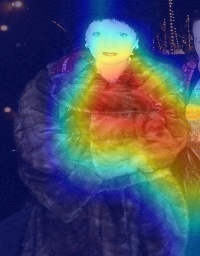} &  \includegraphics[width=24mm,height=24mm]{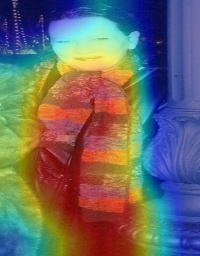} \\
\end{tabular}
\end{adjustbox}
\caption{Grad-Cam analysis of different images within IMAGO, and their respective pictures in IMAGO-FACES and IMAGO-PEOPLE, spreaded over the twentieth century.}
\label{fig:gradcam_dating}
\end{figure*}

Resorting to the Grad-Cam algorithm, it is possible to individuate the cues that led the deep learning models employed in this work to determine the specific year of a photo. From a socio-historical perspective, this analysis may help verifying whether such cues correspond to visual factors which are recognized as representative of a specific year or period.

To this aim, we proceeded analyzing the grad-cam versions of the correctly classified photos, reported in Fig.~\ref{fig:gradcam_dating}, which, for the sake of correctness and simplicity, have all been produced starting from IMAGO dataset images representing two people. In particular, each row corresponds to a specific decade and includes the grad-cam of an IMAGO image in the first column, the grad-cams of the two corresponding IMAGO-FACES images in the second and third columns and, finally, the grad-cams of the two derived IMAGO-PEOPLE images in the fourth and fifth columns, respectively.

From this Figure, it is possible to see that the single-input classifiers (IMAGO, IMAGO-FACES and IMAGO-PEOPLE, respectively), which processed the whole image, its faces and its people's full figures, focused on different regions (red areas in Fig.~\ref{fig:gradcam_dating}). This corroborates the finding that images contain complementary knowledge, located in different segments, that can be exploited to improve the performance of the dating model, as discussed in Section \ref{sec:results_dating}. In other words, not only the faces of the people represented in a given image, but also their full figures and the whole photo can support a more accurate dating. 
This argues for the fact, which is perfectly comprehensible in socio-historical terms, that also clothing and furniture components, as well as their style, play an important role in the temporal classification of a picture.

\begin{figure*}[h]
\centering
\begin{adjustbox}{width=0.85\linewidth}
\begin{tabular}{m{22mm}!{\vrule width 2pt}m{24mm}m{24mm}m{24mm}m{24mm}m{24mm}!{\vrule width 2pt}}
    \multicolumn{1}{c!{\vrule width 2pt}}{} & \multicolumn{5}{c!{\vrule width 2pt}}{\textbf{IMAGO}} \\
    \textbf{Affectivity} & 
    \includegraphics[width=24mm,height=24mm]{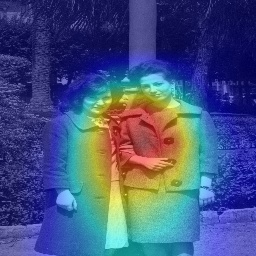} & 
    \includegraphics[width=24mm,height=24mm]{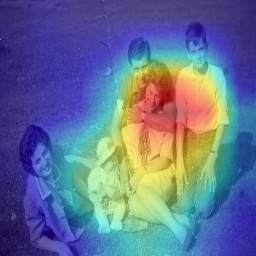} &
    \includegraphics[width=24mm,height=24mm]{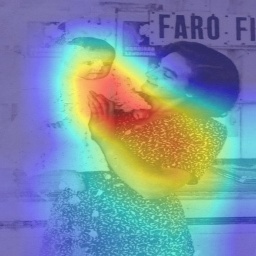} & 
    \includegraphics[width=24mm,height=24mm]{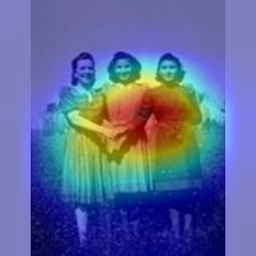} &  
    \includegraphics[width=24mm,height=24mm]{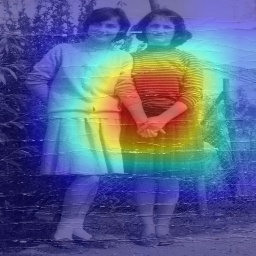} \\
    \textbf{Fashion} & 
    \includegraphics[width=24mm,height=24mm]{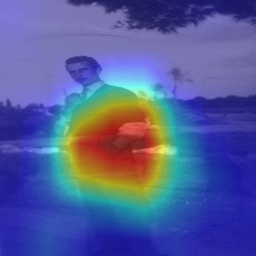} &   \includegraphics[width=24mm,height=24mm]{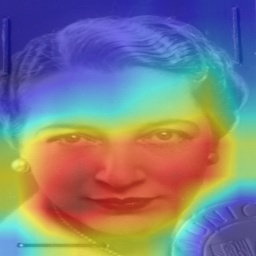} &   \includegraphics[width=24mm,height=24mm]{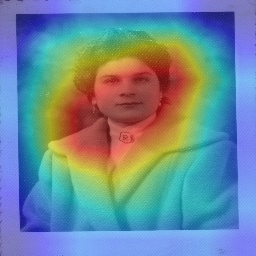} &   \includegraphics[width=24mm,height=24mm]{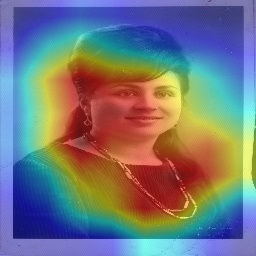} &   \includegraphics[width=24mm,height=24mm]{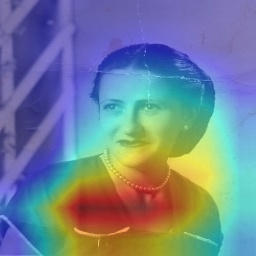} \\
    \textbf{Motorization} & \includegraphics[width=24mm,height=24mm]{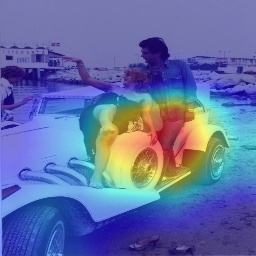} &   \includegraphics[width=24mm,height=24mm]{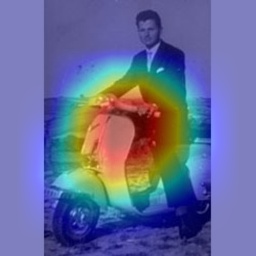} &   \includegraphics[width=24mm,height=24mm]{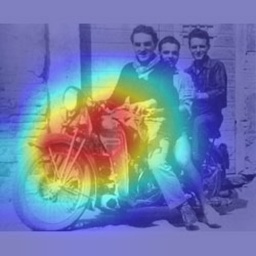}&   \includegraphics[width=24mm,height=24mm]{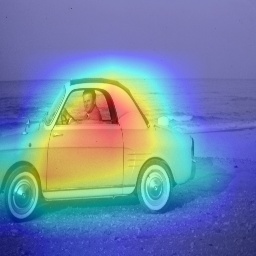}&   \includegraphics[width=24mm,height=24mm]{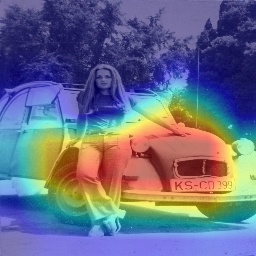} \\
    \textbf{Music} &  \includegraphics[width=24mm,height=24mm]{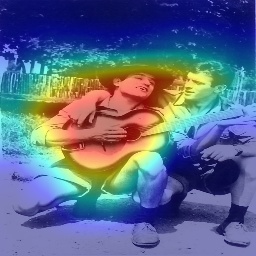} &   \includegraphics[width=24mm,height=24mm]{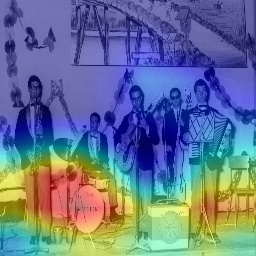} &   \includegraphics[width=24mm,height=24mm]{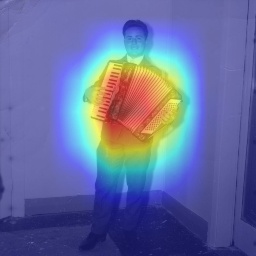}&   \includegraphics[width=24mm,height=24mm]{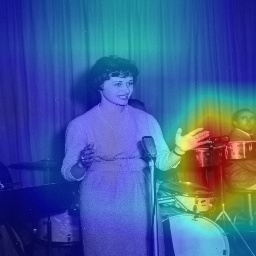}&   \includegraphics[width=24mm,height=24mm]{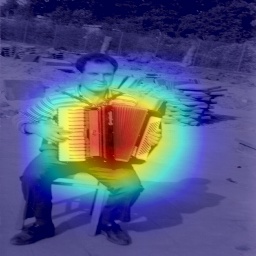} \\
    \textbf{Politics} & \includegraphics[width=24mm,height=24mm]{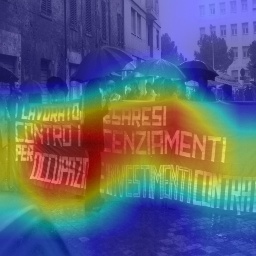} &   \includegraphics[width=24mm,height=24mm]{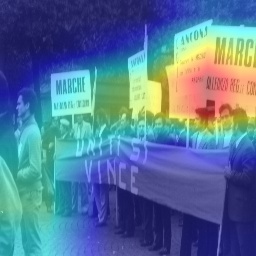} &   \includegraphics[width=24mm,height=24mm]{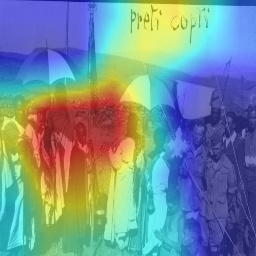}&   \includegraphics[width=24mm,height=24mm]{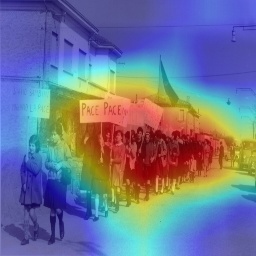}&   \includegraphics[width=24mm,height=24mm]{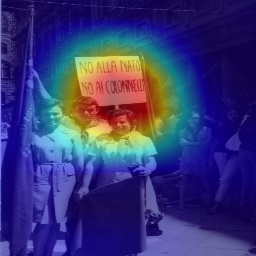} \\
    \textbf{Rites} & \includegraphics[width=24mm,height=24mm]{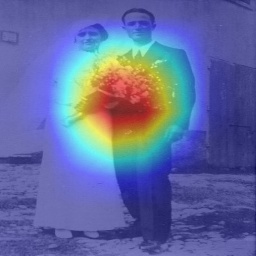} &   \includegraphics[width=24mm,height=24mm]{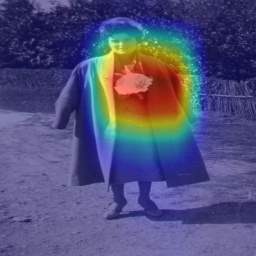} &   \includegraphics[width=24mm,height=24mm]{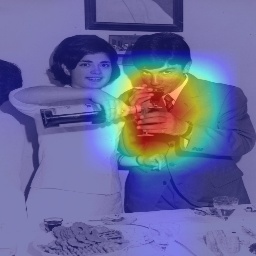}&   \includegraphics[width=24mm,height=24mm]{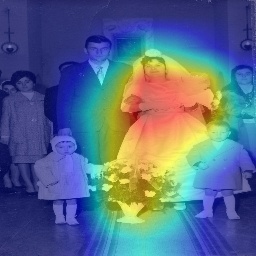}&   \includegraphics[width=24mm,height=24mm]{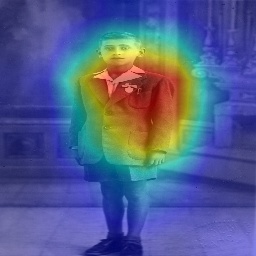}  \\
    \textbf{School} & \includegraphics[width=24mm,height=24mm]{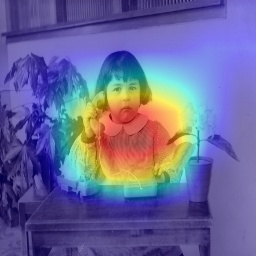} &   \includegraphics[width=24mm,height=24mm]{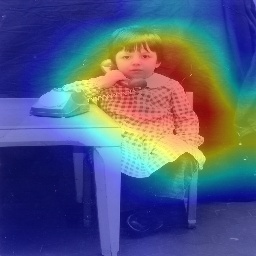} &   \includegraphics[width=24mm,height=24mm]{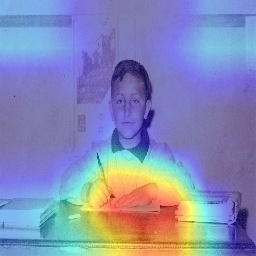}&   \includegraphics[width=24mm,height=24mm]{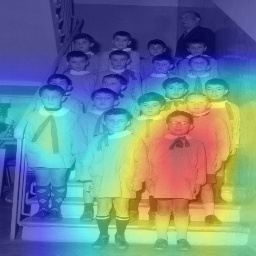}&   \includegraphics[width=24mm,height=24mm]{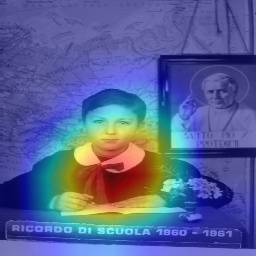} \\
\end{tabular}
\end{adjustbox}
\caption{Grad-Cam analysis of socio-historical context of pictures within IMAGO.}
\label{fig:gradcam_semantic}
\end{figure*}

\begin{figure*}[h]
\centering
\includegraphics[width=.49\linewidth]{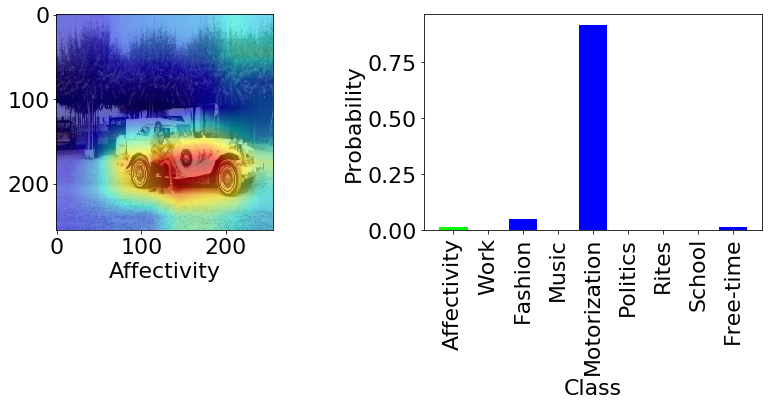}
\includegraphics[width=.49\linewidth]{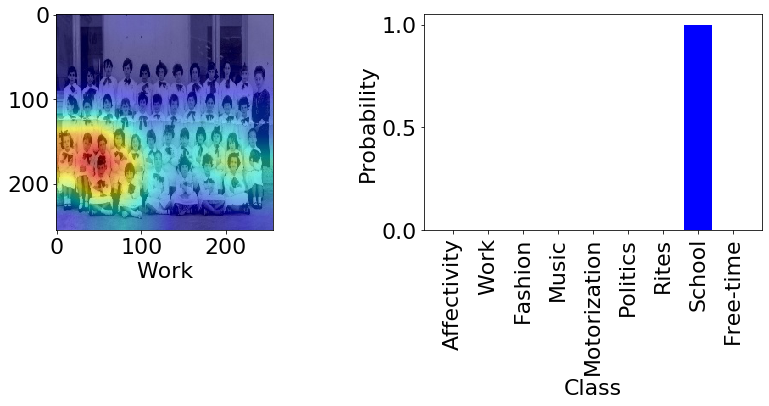}
\caption{Grad-Cam analysis related to wrong classifications: these images exhibit the semantic overlapping of different socio-historical classes (\textit{Affectivity} recognized as \textit{Motorization} and \textit{Work} recognized as \textit{School}).}
\label{fig:wrong_gradcam_classification_1}
\end{figure*}

\begin{figure*}[h]
\centering
\includegraphics[width=.49\linewidth]{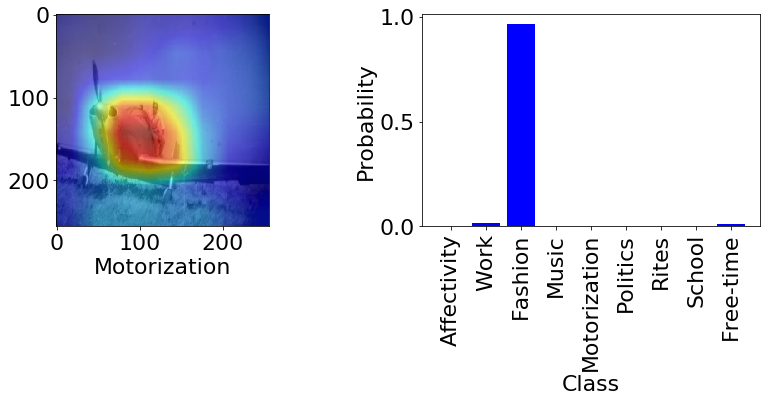}
\includegraphics[width=.49\linewidth]{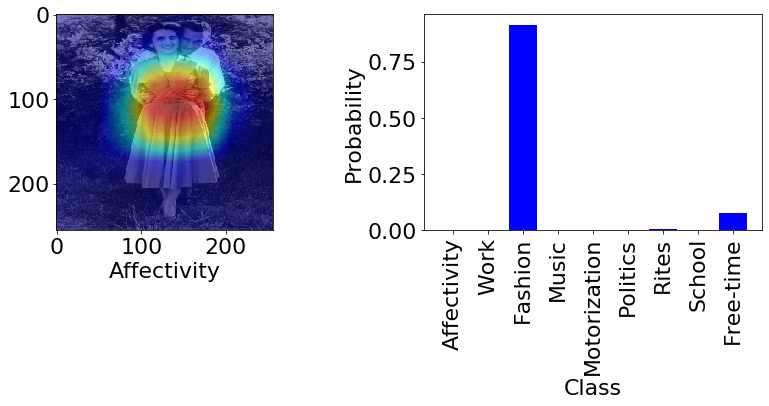}
\caption{Grad-Cam analysis related to wrong classifications: these images exhibit the bias of the neural network for different socio-historical classes (\textit{Motorization} and \textit{Free-time} recognized as \textit{Fashion}).}
\label{fig:wrong_gradcam_classification_2}
\end{figure*}

\subsection{Socio-historical context classification task}
\label{subsec:socio_gradcam}

Considering now the socio-historical context classification task, we show a sample of correctly classified IMAGO images processed by the Grad-Cam algorithm in Fig.~\ref{fig:gradcam_semantic}. Per each row, starting from the top, five exemplary images belonging to the \textit{Affectivity}, \textit{Fashion}, \textit{Motorization}, \textit{Music}, \textit{Politics}, \textit{Rites} and \textit{School} classes are produced. Such images are representative of the concrete and abstract concepts which the IMAGO classifier learned for each class.

More in detail, specific gestures which characterize the affection between people (e.g., hugs, kisses, hold a baby, handshakes) as shown in the first row of Fig.~\ref{fig:gradcam_semantic} are characteristic of the \textit{Affectivity} class.
Specific objects like earrings, necklaces and lapels, but also particular hairstyles, are used to classify a picture as belonging to the \textit{Fashion} class (second row of the Figure).
All kinds of vehicles, as well as musical instruments, are put to good use to recognize a given picture as member of the \textit{Motorization} or the \textit{Music} classes, shown in the third and fourth rows, respectively.
The presence of a political banner is typical of pictures in the \textit{Politics} class (fifth row).
The model distinguishes the objects and gestures (e.g., white dress, flowers, pour a drink, cheers) that characterize the \textit{Rites} class (sixth row). Finally, children wearing school uniforms, as well as school gear (e.g., books, pens, desks) are used to recognize pictures in the \textit{School} class (last row). Please note that the same type of analysis was not possible for the \textit{Free-time} and \textit{Work} classes. In particular, for the former it was not possible to individuate distinguishing cues which clearly helped identifying the images there contained (confirming the quantitative results reported in Section \ref{subsec:result_socio_historical}). The latter class, instead, was difficult to correctly recognize as its cues overlapped with the \textit{Affectivity}, \textit{Fashion} and \textit{Free-time} classes, as exhibited in Fig. \ref{fig:conf_matrix_semantic}.

It is not surprising that the model was able to correctly classify pictures belonging to the \textit{Motorization} and \textit{Music} classes, as these are clearly characterized by specific objects, but, more importantly, already part of the model pre-trained on ImageNet. However, also for the majority of the other classes (not studied so far in literature, to the best of our knowledge), the model is able to isolate and focus on the details which distinguish them. 

However, all that is gold does not glitter. Fig.~\ref{fig:wrong_gradcam_classification_1} reports a few misclassified examples. From the leftmost picture and its probability histogram, it is possible to see that a photo containing a car was classified as belonging to the \textit{Motorization} class, but the ground truth label assigned to the picture was \textit{Affectivity} (two people are hugging each other in front of the car).
Instead, the rightmost picture and its corresponding probability histogram show that a picture depicting a school class was classified as belonging to  \textit{School}, while the actual one was \textit{Work}  (a teacher is standing in the rightmost part of the picture).
Such ambiguities may be traced back to the fact that the IMAGO dataset has been labeled by human classifiers which could adopt different points of view. In fact, the leftmost picture presented in Fig.~\ref{fig:wrong_gradcam_classification_1}, may be classified as belonging to the \textit{Motorization} class or to the \textit{Affectivity} one, without be mistaken in any of the two cases. Likewise, the rightmost picture could be classified as belonging to the \textit{Work} class, when adopting a teacher's point of view, but also to the \textit{School} one, when instead adopting a students' perspective. We are evidently experiencing a well known phenomenon in cataloguing studies, known as the observer's bias \cite{observer_bias}. In such setting, also the deep learning model becomes an observer and as such provides classifications which reflect the knowledge it has acquired. The existence of the observer's bias emphasizes the importance of the Top-$k$ classification (described in Section \ref{subsec:result_socio_historical}) and may open to the possibility of investigating the use of  multiple socio-historical labels for an image.

Not related to the observer's bias, it turns out that there have also been misclassifications due to neural network biases. Two representative examples are reported in Fig.~\ref{fig:wrong_gradcam_classification_2}: both of them were classified belonging to the \textit{Fashion} class but the leftmost picture should belong to \textit{Motorization} while the rightmost to \textit{Affectivity}. 
The classes provided by the model in this case are quite off target, since the leftmost picture depicts a man standing by a plane, while the rightmost contains a couple of people hugging each other. These errors were driven by the fact that the model here focused on secondary details of the pictures (i.e., fashion aspects), rather than on the main ones (i.e., the structure of the plane and the hugging gesture).

\section{Conclusions and future works}
\label{sec:conclusions}
In this paper we introduced the IMAGO dataset, which is composed by pictures belonging to family albums. We trained and tested single and multi-input deep learning models which exploited different regions of a given photo to identify the year in which it was taken. We showed how a higher number of faces and people positively affected the final accuracy of the dating model. Furthermore, we exhibited the existence of complementary information that could be exploited by the model to improve the accuracy of the dating task. 

In addition, to the best of our knowledge, this is the first work to introduce a socio-historical context classification task of family album images. This task consists in identifying the sociological and historical context of a picture, according to the definitions provided by socio-historical scholars~\cite{calanca2011italians}. We observed that the complementary knowledge learned in advance from people's faces and full figures did not help improving the classification accuracy, as relevant regions and their spatial relationships may be located throughout a photo. We also observed that the deep learning models, trained for the socio-historical context classification task, were able to correctly classify abstract concepts (i.e., affectivity, politics, rites) by focusing on relevant and actual cues (e.g, kisses and hugs for the affection, banners for the politics class, marriage cakes and toasts for the rites class).

For the possible developments of this work, we resort to what Antonio Gramsci said: ``The starting-point of critical elaboration is the consciousness of what one really is, and is 'knowing thyself' as a product of the historical processes to date, which has deposited in you an infinity of traces, without leaving an inventory.'' Indeed, when focusing on the image dating or socio-historical classification task, scholars perform analyses which resort at once to different sources of information (e.g., journals, magazines, archival documents), as well as to the traces mentioned by Antonio Gramsci. The same objectives, at the state of the art, are pursued in a completely different way when resorting to computer vision techniques, which are nowadays limited to what can be directly learnt from an image. In essence, historical information cannot be surely drawn when solely analyzing one of the traces that have been left on the ground, but should also consider the influence of space, time and of the human observer. Interestingly, in this work we were able to highlight how such traces also permeated the socio-historical classification. 
Clearly, this is only one step in the direction of creating holistic models which may take into account all the processes involved in the complex socio-historical domain.

Finally, this line of work may benefit from further investigations which included: (i) a larger and more balanced amount of data, (ii) an additional exploitation of the complementarity between the different image regions and (iii) a more thorough analysis of the concept of deep neural networks viewed as observers within the socio-historical context classification task.

\ifCLASSOPTIONcompsoc
  \section*{Acknowledgments}
\else
  \section*{Acknowledgment}
\fi

This work was supported by the University of Bologna with the Alma Attrezzature 2017 grant and by AEFFE S.p.a. and the Golinelli Foundation with the funding of two Ph.D. scholarships.

\ifCLASSOPTIONcaptionsoff
  \newpage
\fi

\bibliographystyle{IEEEtran}
\bibliography{IEEEabrv,bibfile.bib}






\newpage

\end{document}